\documentclass[10pt,journal,compsoc]{IEEEtran}
\usepackage{comment}
\usepackage{enumitem}
\usepackage{multirow}
\usepackage{balance}
\usepackage{booktabs}
\usepackage{ulem} 
\usepackage{graphicx}
\usepackage{subfigure}
\usepackage[colorlinks, linkcolor=blue, anchorcolor=blue, citecolor=blue]{hyperref}

\usepackage[linesnumbered]{algorithm2e}
\usepackage{algpseudocode}  
\usepackage{amsmath}  

\newtheorem{mydef}{Definition}

\usepackage{amsmath,amssymb,amsfonts}

\ifCLASSOPTIONcompsoc
  \usepackage[nocompress]{cite}
\else
  \usepackage{cite}
\fi
\ifCLASSINFOpdf
\else
\fi
\begin{document}
%
\title{Collaborative Graph Neural Networks for Attributed Network Embedding}

\author{Qiaoyu Tan,~\thanks{Qiaoyu Tan is with the Department of Computer Science and Engineering, Texas A\&M University, Texas, USA. E-mail: qytan@tamu.edu.}Xin Zhang, Xiao Huang, Hao Chen,~\thanks{Xin Zhang, Xiao Huang, and Hao Chen are with the Department of Computing, The Hong Kong Polytechnic University, Hung Hom, Hong Kong. E-mail: \
xin12.zhang@connect.polyu.hk; xiaohuang@comp.polyu.edu.hk;
sundaychenhao@gmail.com.}Jundong Li,~\thanks{Jundong Li is with the Department of Electrical and Computer Engineering, University of Virginia, Virginia, USA. E-mail: jundong@virginia.edu.}and Xia Hu~\thanks{Xia Hu is with the Department of Computer Science, Rice University, Texas, USA. E-mail: xia.hu@rice.edu.}









}

%
%

\markboth{Preprint}%
{Shell \MakeLowercase{\textit{et al.}}: Bare Demo of IEEEtran.cls for Computer Society Journals}
%



\IEEEtitleabstractindextext{%
\begin{abstract}
Graph neural networks (GNNs) have shown prominent performance on attributed network embedding. However, existing efforts mainly focus on exploiting network structures, while the exploitation of node attributes is rather limited as they only serve as node features at the initial layer. This simple strategy impedes the potential of node attributes in augmenting node connections, leading to limited receptive field for inactive nodes with few or even no neighbors. Furthermore, the training objectives (i.e., reconstructing network structures) of most GNNs also do not include node attributes, although studies have shown that reconstructing node attributes is beneficial. Thus, it is encouraging to deeply involve node attributes in the key components of GNNs, including
graph convolution operations and training objectives. 
However, this is a nontrivial task since an appropriate way of integration is required to maintain the merits of GNNs.
To bridge the gap, in this paper, we propose COllaborative graph Neural Networks--CONN, a tailored GNN architecture for attribute network embedding. It improves model capacity by 1) selectively diffusing messages from neighboring nodes and involved attribute categories, and 2) jointly reconstructing node-to-node and node-to-attribute-category interactions via cross-correlation. Experiments on real-world networks demonstrate that CONN excels state-of-the-art embedding algorithms with a great margin. 

\end{abstract}

}

\maketitle

\IEEEdisplaynontitleabstractindextext

%
\IEEEpeerreviewmaketitle

\IEEEraisesectionheading{\section{Introduction}\label{sec:introduction}}

%
%
%
%
\IEEEPARstart{A}{ttributed} networks~\cite{huang2017label,gao2018deep,hong2019deep,wang2021decoupled} are ubiquitous in a myriad of real-world information systems, such as academic networks and social medial systems. Unlike plain networks in which only node-to-node interactions are available, each node in attributed network is associated with a rich set of attributes, describing its distinctive characteristics. For example, in social networks, users connect with others as friends, share opinions, and post comments as attributes. In academic citation networks, different articles are connected via citation links, and each article has substantial text information, such as an abstract sentence to describe its own topic. Several studies~\cite{mcpherson2001birds,marsden1993network} in social science have revealed that attributes of nodes can reflect and affect their community structures~\cite{yang2010understanding} in practice. Thus, it is encouraging and important to study attributed networks. To this end, attributed network embedding~\cite{liao2018attributed,huang2017label,wang2017attributed}, targeting at leveraging both network proximity and node attribute affinity to learn low-dimensional node representations, has attracted great attention in recent years, and many efforts have been devoted from both academia and industry~\cite{wang2019learning,yang2018binarized,gao2020community,rozemberczki2021multi,kipf2016variational,hamilton2017inductive,li2017attributed}. 

Among them, embedding paradigm based on graph neural networks (GNNs)~\cite{gilmer2017neural,hamilton2017inductive,han2022geometric,gao2018large,jiang2022fmp,li2021understanding,jiang2022topology,kipf2016semi,tan2023s2gae,jiang2022fair,li2018deeper,yu2022heterogeneous,tan2022kd,miao2021lasagne,tan2019deep,han2022geometric} has achieved remarkable success over a variety of downstream graph analytical tasks, including node classification~\cite{huang2017accelerated,tan2022model,gao2021self}, graph classification~\cite{han2022g,zhang2022graph}, link prediction~\cite{tan2023bring,zhang2018link,zhang2021labeling,tan2020learning}, node clustering~\cite{he2022gnns,fettal2022efficient}, and anomaly detection~\cite{zhou2021subtractive,zhou2022unseen,zhang2023tkde,dong2023active}. The design recipes for GNNs based methods include two major components: 1) a GNN encoder that takes node attributes and node-to-node interaction network as input and outputs low-dimensional node representations;
2) the training objective, which is derived to reconstruct the input data (e.g., network structure), so as to train the model unsupervised.
Thanks to the critical message propagation mechanism in GNNs, the GNN encoder is naturally applicable for attributed networks. Therefore, existing GNNs-based embedding efforts mainly focus on advancing model performance with more expressive message passing schema, such as adaptively aggregating messages from neighbors via the attention layer~\cite{velivckovic2017graph,gao2021higher,luo2020g}


Despite their popularity and recent advances in refining GNNs architectures to effectively model the topological structures~\cite{xu2018powerful,gao2019graph,feng2020graph,chen2020simple,stachenfeld2020graph}, little attention was paid to the node attributes. Prior GNNs studies focus on updating the representation of each node by aggregating the representations of itself and its neighbors~\cite{wu2019simplifying} recursively. 
In this learning process, node attributes are merely employed as the representations of nodes in the initial layer~\cite{dwivedi2020benchmarking}. They would be blocked from message propagation if the network structure is incomplete or missing, which is quite common in real-world applications where graphs exhibit long-tail node degree distribution~\cite{liu2021tail,zheng2021cold}. Besides, even when we design the training objectives of GNNs, the node attributes are seldom used. For instance, the common practice to train GNNs based embedding algorithm for attributed network embedding is to reconstruct the observed node interactions~\cite{hamilton2017inductive}, by either employing a negative sampling based objective or directly recovering the whole input network structure~\cite{kipf2016variational}.
As a summary, node attributes are not well exploited in the existing works.

Recently, there has been a revolution to rethink the value of node attributes in random-walk based embedding approach~\cite{perozzi2014deepwalk,huang2019large}. The core idea is to redesign the crucial component--random-walk in an attribute aware fashion. Specifically, ANRL~\cite{zhang2018anrl} refines the conditional probability that estimates the propensity score between the anchor node and its context by utilizing node attributes. FeatWalk~\cite{huang2019large} conducts an attribute-aware joint random walk to increase the diversity of generated walks. According to their empirical experiments, both of them have shown significant performance gains compared with their vanilla counterparts. Nevertheless, they are tailored for random-walk based approaches and cannot be directly applied to GNN models with trivial efforts. 
Motivated by this, in this paper, we propose to explore whether node attributes can be effectively employed to advance the essential building blocks of GNN architectures (i.e., message aggregation mechanism and training objective).

However, it is a non-trivial and challenging task to integrate node attributes into GNN architectures mainly because of two reasons: (i) GNNs already show promising results by integrating node attributes as the initial node representations. It is difficult to further incorporate node attributes into the key components of GNNs, while maintaining the existing benefits and prominent performance~\cite{wu2019simplifying,klicpera2019predict}; (ii) 
real-world node attributes, such as comments of users, abstracts of papers, and descriptions of products, are distinct from the network topological structures and not in line with the graph convolutional operation in GNNs.
Specifically, the values in node attributes are often multi-categorical or continuous variables, while these in the network structures are binary. They are not compatible with each other. Thus a tailored operation is required to learn from them jointly. 
For example, employing autoencoders as the training objective of GNNs would achieve sub-optimal performance~\cite{kipf2016variational,meng2019co}.

To address the aforementioned challenges, we propose a novel unsupervised representation learning model, dubbed \textbf{CO}llaborative Graph \textbf{N}eural \textbf{N}etwork (CONN). It aims to develop a tailored GNN architecture for attributed networks, such that node attributes can be explicitly fused into the message aggregation process as well as the training objective. 
Specifically, we aim to investigate two important research questions. (i) How to leverage node attributes to explicitly guide the message propagation of vanilla GNN, and conduct a collaborative aggregation mechanism? (ii) In the training objective, how to effectively model the heterogeneous interactions, so as to jointly reconstruct the network structure and node attributes? We summarize our major contributions as follows.
 
\begin{itemize}
\item We focus on unsupervised representation learning on attributed networks and propose an effective GNN framework - CONN, to leverage node attributes in the two aforementioned key components of vanilla GNNs.

\item By conducting a bipartite graph on node attributes, we develop a collaborative aggregation mechanism for node embedding. It not only helps to enrich or rebuild node connections through attribute category, but also provides a principled way to update node representation from both 
neighboring nodes and involved attribute categories. 

\item Based on the node and attribute category representations, we design a novel cross-correlation layer to effectively model the complex node or node attribute category interactions. It highlights the similarity of two anchor nodes from their multi-granularity features and significantly boosts the reconstruction capability of vanilla GNNs. 

\item We evaluate CONN on node classification and link prediction tasks. Empirical results on benchmark datasets show that CONN performs consistently better than other state-of-the-art embedding methods. Moreover, we also analyze the robustness and convergence speed of CONN in Section~\ref{sec:further}.
\end{itemize}

 \begin{figure*}[h]
  \centering
  \includegraphics[width=16.8 cm]{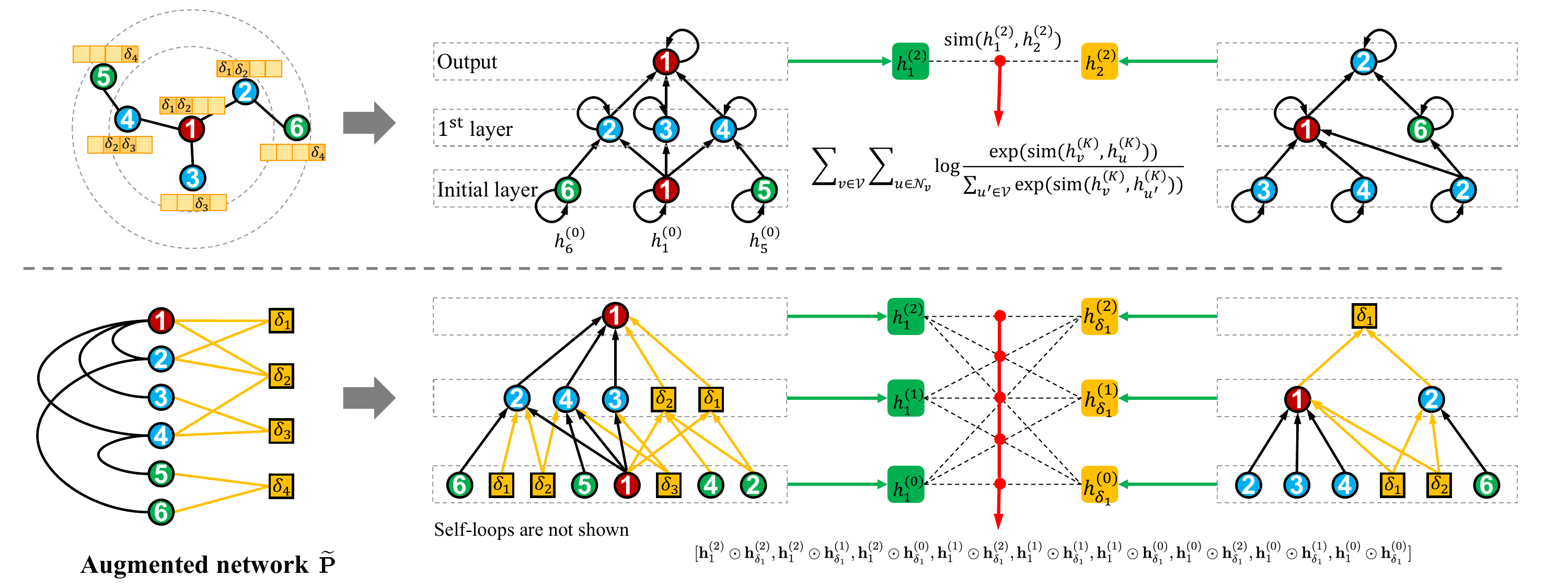}
  \vspace{-6pt}
  \caption{Instead of inferring the node-to-node links by using node attributes as side features, CONN targets at jointly learning the network structure and node attributes in a unified latent space under graph neural network framework.
  }
  \label{figure_flowchart}
\end{figure*}

 

\section{Related Work}\label{sec:relatework}
There are three types of related works, based on whether node attributes are modeled for network embedding or not. In this section, we briefly
review some related works in these two fields. Please refer to~\cite{zhang2020deep,zhou2018graph,wu2020comprehensive,makarov2021survey,zhang2018network} for comprehensive review. 

\subsection{Network embedding}
The first class of approach can be tracked back to traditional graph machine learning problem, which aims to learn node embedding while preserving the local manifold structure, such as the LPP~\cite{he2003locality} and Laplacian Eigenmaps~\cite{belkin2001laplacian}. Nevertheless, these methods suffer from scalability challenges to large-scale networks, due to the time-expensive eigendecomposition operation on the adjacency matrix, whose time complexity is $O(n^3)$ with $n$ is the number of nodes. Inspired by the recent breakthrough of distributed representation learning~\cite{mikolov2013distributed} in natural language processing, a lot of scalable network embedding methods~\cite{perozzi2014deepwalk,tang2015line,cao2015grarep,grover2016node2vec,wang2016structural,rossi2018deep,tu2018unified,liu2019single} have been developed.
For example, DeepWalk~\cite{perozzi2014deepwalk} and Node2vec~\cite{grover2016node2vec} conduct truncated random walks on the network to generate node sequences, and then feed them into into the Skip-Gram algorithm~\cite{mikolov2013distributed} for node embedding. LINE~\cite{tang2015line} optimizes the first and second order neighborhood associations to learn representations.  
GraRep~\cite{cao2015grarep} extends LINE to capture high-order neighborhood relationships.
SDNE~\cite{wang2016structural} applies deep learning for node embedding and targets to capture the non-linear graph structure as well as preserve the global and local structures of the graph.~\cite{wang2017community} attempts to incorporate the community structure of the graph for representation learning. {Some other studies aim to deal with large-scale graphs~\cite{yang2019homogeneous,qiu2019netsmf,lin2021large}.}
Despite their simplicity, the aforementioned methods may be limited in practice, since they cannot exploit side information, such as user profiles, worlds in posts, and contexts in photos on social media. 

\subsection{Attributed Network Embedding}
Different from traditional network embedding, attributed network embedding~\cite{zhao2021hierarchical,huang2017accelerated,yang2015network,wang2021hgate,zhao2022multi,zhang2019attributed,liu2021motif,huang2020biane}, which aims to learn node representations by leveraging both the network structure and node attributes, has attracted substantial attention in recent years. For instance, 
ANRL~\cite{zhang2018anrl} incorporates the node attributes into the conditional probability function, which predicts the propensity scores between an anchor node and its context, of Skip-gram to capture structure correlation. {MUSAE~\cite{rozemberczki2021multi} advances Skip-gram model by considering multi-scale neighborhood relationships based on node attributes.} FeatWalk~\cite{huang2019large} aims to conduct attribute-aware random walks to increase the diversity of generated walks, so as to boost the Skip-gram model. {PANE~\cite{yang2020scaling} is another random walk-based method for scalable  training.} TADW~\cite{yang2015network} incorporates node attributes into DeepWalk under matrix factorization framework. PTE~\cite{tang2015pte} employs different orders of world co-occurrence relationships and node label to generate predictive text representations. {ProGAN~\cite{gao2019progan} aims to preserve the underlying proximities  in the hidden space based on a generative adversarial network.} Although the aforementioned methods are capable of employing the network structure and node attributes for joint representation learning, they are limited in capturing the structure information. 

More recently, graph neural networks~\cite{kipf2016semi} based attributed network embedding has attracted increasing attention, due to its ability in capturing structure information, incorporating node attributes, and modeling non-linear relationships. As a pioneering GNN work, GCN~\cite{kipf2016semi} formally suggests to update node presentation by recursively aggregating representations from its adjacent neighbors, and achieves significant performance improvement on semi-supervised classification task. To follow up, GAE~\cite{kipf2016variational} makes the first effort to extend GCN to an autoencoder framework to learn node representations unsupervisedly. Meanwhile, GraphSage~\cite{hamilton2017inductive} improves the scalability of GCN by sampling and aggregating features from a node's local neighborhood. Their results show that GCN is naturally suitable for attributed networks, since it can directly utilize node attributes as initial node features in the first layer for training. Motivated by this, many follow-up studies~\cite{velivckovic2017graph,gao2019graph,liu2020towards} have been proposed to advance model capability by developing more expressive GNN architectures. Recently, some efforts have also been devoted to redefine the training objective of GCN to learn more effective node representations. CAN~\cite{meng2019co} and~\cite{fang2021hyperspherical} advocate to jointly reconstruct the network structure and node attributes under variational autoencoder framework. DGI~\cite{velickovic2019deep} and GIC~\cite{mavromatis2021graph} suggests to learn node representations by maximizing the mutual information between local node representation and the graph summary representation.
GCA~\cite{zhu2021graph} and MVGRL~\cite{hassani2020contrastive} seek to update node representations by maximizing the agreement of representations between different views generated by data augmentation. However, they can only explore the node attributes implicitly in the message propagation process of GNN. In this paper, we propose a principled GNN variant for attributed network embedding, which allows node attributes to explicitly guide the message propagation and training objectives. 

\subsection{Attributed Network Embedding with Extra Knowledge}
{In addition to plain attributed networks, including node features and homogeneous graph structure, some studies also utilize extra knowledge, {such as knowledge graph (KG)~\cite{park2019estimating,zhang2020relational,yasunaga2021qa,dong2023hierarchy},}, to boost the model's performance. For example, PGE~\cite{hou2019representation} studies how to incorporate additional edge features. KGCN~\cite{wang2019knowledge} and KGAT~\cite{wang2019kgat} investigate how to leverage external knowledge graphs (e.g., item knowledge graphs) to improve recommendation quality. However, these methods require additional efforts to generate high-quality information sources, such as edge features and knowledge graphs. In contrast, in this work, we focus on standard attributed network embedding, modeling on homogeneous graphs with pure node attributes, such as continuous features and categorical features.   }


\section{Problem Statement}\label{sec:prelim}
We assume that an attributed network denoted by $\mathcal{G}=(\mathcal{V}, \mathbf{A}, \mathbf{X})$ is given. It has $n$ vertexes collected in set $\mathcal{V}$. These $n$ nodes are connected by an undirected network with its adjacency matrix denoted as $\mathbf{A} \in\mathbb{R}^{n\times n}$. If there is an edge between nodes $v_i$ and $v_j$, then ${\bf A}_{ij}=1$; otherwise, ${\bf A}_{ij}=0$. Besides the network $\mathbf{A}$, each node $v$ also has a descriptive feature vector $\mathbf{x}_v\in\mathbb{R}^m$, known as node attributes, where $m$ is the total number of attribute categories. We denote the neighbor set of node $v$ as $\mathcal{N}_v$, and represent the neighbor set of attribute category $\delta_j$ as ${\mathcal{N}}_{\delta_j}$, i.e., ${\mathcal{N}}_{\delta_j}=\{u| \mathbf{X}_{uj}>0, \text{ for } u\in\mathcal{V}\}$. We employ a diagonal matrix $\mathbf{D}=\text{diag}(d_1,\cdots,d_n)$ to denote the degree matrix, where $d_i=\sum_{j}\mathbf{A}_{ij}$. The main symbols are listed in Table~\ref{tab:notation}.
To study the GNNs in an unsupervised setting, we follow the literature~\cite{kipf2016semi,yao2019graph} and formally define the problem of attributed network embedding in Definition~\ref{define_ane}.

\begin{table}[!t]
  \caption{Summary of notations in this paper.}
    \begin{center}
  \label{tab:notation}
\begin{tabular}{c|l}
\midrule
\textbf{Notation} & \textbf{Description}  \\ 
\midrule
$\mathbf{A}\in\mathbb{R}^{n\times n}$   & adjacency matrix\\
$\mathbf{X}\in\mathbb{R}_+^{n\times{m}}$   & a matrix collects all node attributes\\
$\mathcal{V}$   & a set collects all $n$ nodes\\
$\mathcal{U}$   & a set collects all $m$ attribute categories\\
$\delta_j\in\mathcal{U}$   & the $j^\text{th}$ node attribute category\\
$\mathcal{N}_v$   & a set collects adjacent neighbors of $v$\\
$K$   & the number of graph convolutional layers\\
$\widetilde{\mathbf{P}}\in\mathbb{R}^{(n+m)\times(n+m)}$   & adjacency matrix of augmented network\\
$\mathbf{h}_v^{(k)}\in\mathbb{R}^{1\times d}$   & representation of node $v$ at the $k^\text{th}$ layer\\
$\mathbf{h}_{\delta_j}^{(k)}\in\mathbb{R}^{1\times d}$   & representation of $\delta_j$ at the $k^\text{th}$ layer\\
\toprule
\end{tabular}
\end{center}
\end{table}

\begin{mydef}
{\bf Attributed Network Embedding.} Given an attributed network $\mathcal{G}=(\mathcal{V},  \mathbf{A}, \mathbf{X})$, the goal is to learn a $d$-dimensional continuous vector $\mathbf{h}_v\in\mathbb{R}^d$ for each node $v\in\mathcal{V}$, such that the topological structures in $\mathbf{A}$ and side information characterized by $\mathbf{X}$ could be preserved in the embedding representations $\mathbf{H}$. The performance of this learning task is evaluated by applying $\mathbf{H}$ to various downstream tasks such as node classification and link prediction.
\label{define_ane}
\end{mydef}

To perform attributed network embedding, GNN models~\cite{kipf2016semi,hamilton2017inductive} learn the embedding representation of each node $v$ by aggregating the representations of itself and its neighbors $\mathcal{N}_v$ in the previous layer. A typical neighborhood aggregation mechanism~\cite{kipf2016semi} is expressed as below,
\begin{equation}
\begin{aligned}
\mathbf{h}^{(k)}_v&=\frac{1}{1+d_v}\mathbf{h}_v^{(k-1)} + \sum\nolimits_{u\in\mathcal{N}_v}\frac{\mathbf{A}_{vu}}{\sqrt{(1+d_v)(1 + d_u)}}\mathbf{h}_u^{(k-1)},
\end{aligned}
\label{eq_gcn}
\end{equation}
where $\mathbf{h}_v^{(k)}$ denotes the hidden representation of node $v$ at the $k^{\text{th}}$ layer of GNNs, and $\mathbf{h}_v^{(0)}=\mathbf{x}_v$. The final output is $\mathbf{h}_v^{(K)}$, where $K$ is the maximum number of layers considered. The top subfigure in Figure~\ref{figure_flowchart} illustrates this traditional neighborhood aggregation by using a toy example. We could see that, by stacking two layers, the representations of two-hop neighbors $5$ and $6$ could be accessed by node $v=1$. To train this model in an unsupervised manner, a widely-adopted~\cite{hamilton2017inductive} graph-based loss function is defined as,
\begin{equation}
\begin{aligned}
\mathcal{L}=-\sum\nolimits_{v\in\mathcal{V}}\sum\nolimits_{u\in\mathcal{N}_v}\log\frac{\exp(\text{sim}(\mathbf{h}_v^{(K)}, \mathbf{h}_u^{(K)}))}{\sum_{u'\in\mathcal{V}}\exp(\text{sim}(\mathbf{h}_v^{(K)}, \mathbf{h}_{u'}^{(K)}))}.
\end{aligned}
\label{eq_loss}
\end{equation}
 $\text{sim}(\cdot,\cdot)$ is a similarity function, e.g., inner product. The goal is to make the representations of connected nodes similar to each other, while enforcing the representations of unconnected nodes to be distinct. We observe that, node attributes are employed only as the initial representations ${\bf h}_v^{(0)}$. They have not been further exploited, especially being integrated into the core mechanisms of GNNs.

\section{Collaborative Graph Convolution}\label{sec:prelim}
Node attributes are informative, and significantly correlated with and complementary to the network~\cite{zhang2018anrl,park2020unsupervised,cui2020adaptive,chen2021learning,zhou2023opengsl}. Since they have not been fully exploited in GNNs, we explore to deeply integrate node attributes into the core mechanisms of GNNs, and develop a novel framework named COllaborative graph Neural Network (CONN). Figure~\ref{figure_flowchart} depicts the two major components of CONN, i.e., collaborative neighborhood aggregation and collaborative training objective. First, we redefine the graph convolutions by considering node attribute categories $\mathcal{U}$ as another set of nodes. As illustrated in Figure~\ref{figure_flowchart}, we augment the original network ${\bf A}$ to a new one $\widetilde{\mathbf{P}}$ with $n+m = 6+4$ nodes. It preserves all edges in ${\bf A}$, and contains a link from nodes $v$ to $\delta_j$ if $v$ has a non-zero value in its node attributes ${\bf X}_{vj}$. Notice that $\mathbf{X}_{vj}$ can be both positive or negative, where negative value means the neighbor has negative impact towards the anchor node. Based on $\widetilde{\mathbf{P}}$, we perform neighborhood aggregation. For example, the first-order neighbors of node $1$ have been augmented from $\{2,3,4\}$ to $\{2,3,4,\delta_1, \delta_2\}$, while the second-order neighbors of node $1$ have been augmented from $\{1, 5,6\}$ to $\{1,2,4,5,6,\delta_1, \delta_2,\delta_3\}$. We observe that our collaborative neighborhood aggregation could not only capture node-to-node interactions, but also node-to-attribute-category interactions. Second, to train our model, we design a collaborative loss. The goal is to collaboratively predict all links in the augmented network $\widetilde{\mathbf{P}}$, which incorporates ${\bf A}$ and {\bf X}. Additionally, we design a novel cross correlation mechanism to model the complex interactions between any pair of nodes in $\widetilde{\mathbf{P}}$ (e.g., $1\in\mathcal{V}$ and $\delta_1\in\mathcal{U}$ in Figure~\ref{figure_flowchart}). It employs not only the node representations in the last layer $\mathbf{h}_v^{(K)}$, but also all the remaining ones $\{\mathbf{h}_v^{(k)}, \text{ for } k=0, 1, \ldots, K-1\}$. We now introduce the details in the following subsections.

\subsection{Collaborative Neighborhood Aggregation}
Given an attributed graph $\mathcal{G}=(\mathcal{V},\mathbf{A},\mathbf{X})$, existing GCN architectures mainly define the multiple-hop neighbors of node $v\in\mathcal{V}$ purely based on the network structure $\mathbf{A}$. The first-order neighbors of node $v$ is represented by $\mathcal{N}_{v}=\{u|\mathbf{A}_{uv}=1\}$, while the second-order neighbors is denoted by $\mathcal{N}_{v}^{(2)}=\{u| \mathbf{A}_{uv}^2>0\}$, so on and so forth. $\mathbf{A}^k$ indicates the $k^{\text{th}}$ power of $\mathbf{A}$. As a result, other than serving as the initial node representations, i.e., $\mathbf{h}_v^{(0)}=\mathbf{x}_v$, node attributes are excluded from the graph convolutions, i.e., the core operation of GNNs. As discussed in Introduction section, this could be suboptimal for inactive nodes with few or no neighbors in practice, since they don't have sufficient neighborhood information for GNNs to effectively learn their embedding representations~\cite{liu2021tail}.   


\subsubsection{Augmented network}
\label{augmented_network}
To tackle the aforementioned issue, we propose to leverage the geometrical property of node attributes. We regard each node attribute category $\delta_j\in\mathcal{U}$ as a new node and node attributes ${\bf X}$ as a weighted bipartite graph. By adding it into the original network ${\bf A}$, we would get an augmented network, denoted as $\mathbf{P}$. Mathematically, its adjacency matrix is written as,
\begin{equation}
\begin{aligned}
\mathbf{P} = \left[ 
\begin{array}{cc}
\mathbf{A} & {\mathbf{X}}\\
{\mathbf{X}^\top} & \mathbf{0} 
\end{array} 
\right] \in\mathbb{R}^{(n+m)\times (n+m)}.
\end{aligned}
 \label{eq3}
\end{equation}
{Eq.\eqref{eq3} is applicable for both categorical and continuous node attributes. For continuous features, however, it would generate a dense bipartite graph based on $\mathbf{X}$, which may significantly increase the computation costs when performing convolution on it. To save the computation, we empirically simplify the dense bipartite graph into a sparse one by only preserving the top-$N$ values in each row of $\mathbf{X}$. An appropriate $N$ value acts as a trade-off between the efficiency and accuracy, we analyze its impact in Section~\ref{sec:further}. In summary, we directly use the feature matrix $\mathbf{X}$ of categorical attributes to construct the augmented graph, while {adopting} top-$N$ values in each row of $\mathbf{X}$ to generate a sparse graph for continuous features. }



\subsubsection{Augmented multi-hop neighbors} 
By using $\mathbf{P}$, the first-order and second-order neighbors of node $v$ can be expanded as:
\begin{equation}
\begin{aligned}
\mathcal{N}_{v}&=\{u|\mathbf{A}_{vu}=1\} + \{\delta_j|{\mathbf{X}}_{vj}\neq0\},\\
\mathcal{N}^{(2)}_{v}&=\{u|\mathbf{A}_{vu}^2>0 \ \text{or} \ ({\mathbf{X}}{\mathbf{X}}^\top)_{vu}>0 \} + \{\delta_j|{({\bf A}\mathbf{X}})_{vj}>0\},
\end{aligned}
 \label{eq1}
\end{equation}
where $({\mathbf{X}}{\mathbf{X}}^\top)\in\mathbb{R}^{n\times n}$ collects all node pairs that share at least one node attribute category, i.e., all $v\rightarrow{\delta_j}\rightarrow{u}$ paths. $({\bf A}\mathbf{X})\in\mathbb{R}^{n\times m}$ implies node-to-attribute-category interactions reflected by $v\rightarrow{u}\rightarrow{\delta_j}$ paths.
Compared with traditional GNNs, we explicitly model the node-to-attribute-category interactions within $K$ hops. Original node-to-node interactions have been enriched by  paths passing through $\delta_j\in\mathcal{U}$.
Similarly, the first and second-order neighbors of $\delta_j\in\mathcal{U}$ could be computed as,
\begin{equation}
\begin{aligned}
\mathcal{N}_{\delta_j}&=\{u|{\mathbf{X}}_{uj}\neq0\}, \\ 
\mathcal{N}^{(2)}_{\delta_j}&=\{u|{(\mathbf{A}\mathbf{X})}_{uj}>0\} + \{\delta_i|{(\mathbf{X}^\top{\bf X})}_{ji}>=1\},
\end{aligned}
 \label{eq2}
\end{equation}
where $(\mathbf{X}^\top{\bf X})\in\mathbb{R}^{m\times m}$ denotes the attribute category correlations estimated by their common nodes. It collects $\delta_j\rightarrow{u}\rightarrow{\delta_i}$ paths. We enable nodes in $\mathcal{V}$ to propagate messages to attribute categories, and correlations among attribute categories to affect the graph convolution. 
Based on this, we can not only enrich node interactions using attribute-category as intermediate to improve model performance (see Table~\ref{table_classify} and~\ref{table_lp}), but also enhance model robustness \textit{w.r.t.} missing edges as shown in Section~\ref{sec:further}.

Another thing we want to remark is that different from previous efforts~\cite{wang2020gcn,liu2021self} that define a node-to-node similarity network based on feature distance, we directly build a node-to-attribute-category bipartite graph on feature matrix by using attribute values as edge weights. As analyzed before, a bipartite graph between node and attribute-category can not only enrich or rebuild node interactions using attribute as intermedia, but also preserve feature information as much as possible.  

\subsubsection{Collaborative aggregation}

We now illustrate how to learn node representations based on the augmented network ${\bf P}$. Given the recent advances in heterogeneous graph embedding~\cite{zhang2019heterogeneous,zhang2019heterogeneous,jin2021heterogeneous}, the intuitive solution is to apply these well-established heterogeneous GNNs to learn embeddings from $\mathbf{P}$ (it consists of two objects (node and attribute-category) and two relations (node-to-node and node-to-attribute-category)). However, our preliminary experiments show that these methods perform not good on our \textit{synthetic heterogeneous graph} (See discussion in~\ref{heterogenous_varaint}), since they may over-emphasize the heterogeneity between node and its attributes, making the learning process substantially complex. Since standard GNNs architectures~\cite{velivckovic2017graph,gao2019graph,liu2020towards,meng2019co} could already achieve high performance on attributed networks by looking them as homogeneous graph, we propose to follow the tradition and regard the augmented network $\bf P$ as homogeneous resource with simple weight schema. We leave attributed network embedding from heterogeneous GNNs perspective as our future work. 

Specifically, our essential idea is to treat node vertex and attribute-category vertex as identical vertices but do provide a weight hyperparameter $\alpha $ to control the information diffusion between the network $\bf A$ and node attributes $\bf X$. 
It is because the importance of node-to-node interactions and node attributes are not explicitly available. In our setting, the binary values in ${\bf A}$ might not be compatible with the feature values in ${\bf X}$ (e.g., continuous features), we define a refined transition probability matrix as follow.     
\begin{equation}
\begin{aligned}
\widetilde{\mathbf{P}} = \left[ 
\begin{array}{cc}
\alpha\widetilde{\mathbf{A}} &(1-\alpha)\widetilde{\mathbf{X}}\\
(1-\alpha)\widetilde{\mathbf{X}}^\top & \alpha\mathbf{I} 
\end{array} 
\right],
\end{aligned}
 \label{eq5}
\end{equation}
where $\widetilde{\mathbf{A}}$ and $\widetilde{\mathbf{X}}$ denote the normalization of $(\mathbf{A}+{\bf I})$ and $\mathbf{X}$ after applying $\ell_1$ norm to normalize each row respectively. $\alpha\in[0,1]$ is a trade-off hyper-parameter to impose our inductive bias about the importance of network structure and node attributes. Specifically, when $\alpha=1$, it yields to a vanilla graph convolutional operation purely based on the network structure. As $\alpha$ increases, node representations would be more dependent on node attributes, and node-to-attribute-category interactions and attribute category correlations will be gradually incorporated into the graph convolution process. 


Based on $\widetilde{\mathbf{P}}$, standard graph convolutional layers can be directly applied to update node representations. We follow a simple GNN model~\cite{wu2019simplifying}, and update the corresponding embedding matrix with a simple sparse matrix multiplication. We use $\mathbf{H}^{(k)}\in\mathbb{R}^{(n+m)\times d}$ to denote the intermediate representation of all $(n+m)$ nodes in the $k^{\text{th}}$ layer. Mathematically, it could be written as,
\begin{equation}
\mathbf{H}^{(k)}\leftarrow\widetilde{{\bf P}}^K\mathbf{H}^{(0)}.
 \label{eq4}
\end{equation}
Our initial node representations  $\mathbf{H}^{(0)}\in\mathbb{R}^{(n+m)\times d}$ are not based on ${\bf X}$. Instead, $\mathbf{H}^{(0)}$ is a trainable embedding matrix that is randomly initialized following common protocols~\cite{perozzi2014deepwalk,qiu2018network}. To further illustrate the correlation between $\mathbf{h}^{(k)}_v$ and $\{\alpha, {\bf A}, {\bf X}\}$, we rewrite the corresponding update rule as follows,
\begin{equation}
\begin{aligned}
\mathbf{h}^{(k)}_v=&\alpha\widetilde{\mathbf{A}}_{vv}\mathbf{h}_v^{(k-1)} + \alpha\sum\nolimits_{u\in\mathcal{N}_v}\widetilde{\mathbf{A}}_{vu}\mathbf{h}_u^{(k-1)}\\ &+ (1-\alpha)\sum\nolimits_{\delta_j\in\mathcal{N}_v}\widetilde{\mathbf{X}}_{vj}\mathbf{h}_{\delta_j}^{(k-1)}.
\end{aligned}
\label{eq_conn}
\end{equation}
Eq.~\eqref{eq_conn} provides a principled solution to utilize and control node attributes. On the one hand, it can explicitly enrich or replenish node interactions by treating attribute-category as additional node. On the other hand, neighborhood information from node and attribute-category are selectively combined via trade-off parameter $\alpha$. 


\subsection{Collaborative Training Objective}
Given the updated representations $\mathbf{H}^{(k)}$ for $k=0,1,\ldots,K$, we need an unsupervised objective to train the GNN model. A widely-adopted approach~\cite{kipf2016variational,hamilton2017inductive,meng2019co} is to employ the node representations at the last layer $\mathbf{H}^{(K)}$ to reconstruct all edges. As illustrated in Eq.~\eqref{eq_loss}, it estimates the probability of two nodes $v$ and $u$ being connected based on the similarity between their vectors $\mathbf{h}_v^{(K)}$ and $\mathbf{h}_u^{(K)}$. This approach has been demonstrated to be effective in plain networks, but node attributes are often available in practice. Eq.~\eqref{eq_loss} could not directly incorporate node attributes. Another intuitive solution~\cite{kipf2016variational} is to employ autoencoders to reconstruct both ${\bf A}$ and ${\bf X}$. It would achieve suboptimal performance because the topological structures ${\bf X}$ are heterogeneous with node attributes ${\bf A}$. 

\subsubsection{Cross correlation mechanism}
To cope with aforementioned issues, we propose a novel cross correlation mechanism to model and predict the complex node-to-node and node-to-attribute-category interactions. It has two major steps.
First, we remove all weights in ${\bf P}$ and convert it into a binary matrix. We target at integrating the network and node attribute into our training objective. To make them compatible with each other, we define a binary adjacency matrix as, 
\begin{equation}
\begin{aligned}
\mathbf{P}^* = \left[ 
\begin{array}{cc}
\mathbf{A} & {\mathbf{X}^*}\\
{{\mathbf{X}^*}^\top} & \mathbf{0} 
\end{array} 
\right] \in\mathbb{R}^{(n+m)\times (n+m)},
\end{aligned}
 \label{eq3}
\end{equation}
where $\mathbf{X}^*_{v\delta_j} =1$ if $\mathbf{X}_{v\delta_j} > 0$. Our goal is to recover the node-to-node and node-to-attribute-category interactions in $\mathbf{P}^*$.
Second, the node representations $\mathbf{H}^{(k)}$ for $k=0,1,\ldots,K$ at all layers are learned based on different orders of neighbors. One merit of our model is that we have integrated node attributes ${\bf X}$ into the collaborative neighborhood aggregation and no longer need to employ ${\bf X}$ as the initial node representations $\mathbf{H}^{(0)}$. Since we are flexible to define $\mathbf{H}^{(0)}$, we could make the dimensions of all $\{\mathbf{H}^{(k)}\}$ the same. In such a way, we could easily take full advantage of them, and model the complex interaction from node $v$ to node $u$ or node $v$ to node attribute category $\delta_j$ as,
\begin{equation}
\begin{aligned}
\mathbf{y}_{vu} = \text{MLP}(||_{k=0}^K||_{i=0}^K\mathbf{h}_{v}^{(k)}\odot\mathbf{h}_{u}^{(i)}), \\ \mathbf{y}_{v\delta_j} = \text{MLP}(||_{k=0}^K||_{i=0}^K\mathbf{h}_{v}^{(k)}\odot\mathbf{h}_{\delta_j}^{(i)})
\end{aligned}
 \label{node2node}
\end{equation}
where $||$ indicates the concatenation operation. $\mathbf{h}_{v}^{(k)}\odot\mathbf{h}_{u}^{(i)}\in\mathbb{R}^{n\times d}$ denotes the correlation feature between the under-$(k+1)^{\text{th}}$-order neighborhood of node $v$ and the under-$(i+1)^{\text{th}}$-order neighborhood of node $u$. $\odot$ represents the element-wise multiplication. $\mathbf{y}_{vu}\in\mathbb{R}$ and $\mathbf{y}_{v\delta_j}\in\mathbb{R}$ are the predicted scores for pairs $(v,u)$ and $(v,\delta_j)$, respectively. MLP denotes a three-layer multilayer perceptron with a Relu activation function. 

{It is worth noting that the element-wise operation between node representations has been explored in KGAT~\cite{wang2019kgat} and NGCF~\cite{wang2019neural} for the recommendation. Our method differs in two ways. First, KGAT and NGCF utilize this technique to facilitate message propagation in each GNN layer (for node-level embedding). Yet, CONN applies it to generate edge representations of two end nodes from different granularities. Second, the proposed cross-correlation mechanism is far more element-wise. Its novelty lies in the proposal to obtain an informative edge representation of end nodes by integrating their cross-correlations from different combinations of the GNN layers. Therefore, the proposed cross-correlation layer is different from KGAT and NGCF because it aims to improve the quality of edge representation while the referenced methods work on boosting the representations of each node per GNN layer, from which they are orthogonal to us and can be incorporated as the base GNN backbone.}

\subsubsection{Analysis}
The subfigure on the bottom of Figure~\ref{figure_flowchart} illustrates the key idea of the proposed cross correlation mechanism. Given any two entities (e.g., node $v=1$ and $\delta_1$ in Figure~\ref{figure_flowchart}), we aim to capture the second-order correlations across all $K+1$ embedding representations of the two entities. So, in total, Eq.~\eqref{node2node} has concatenated $(K+1)^2$ correlation features (e.g., $9$ in Figure~\ref{figure_flowchart}), which is a small number. It should be noted that the element-wise multiplication~\cite{zhu2020bilinear} would highlight the shared patterns within the two input node representations, e.g., $\mathbf{h}_{v}^{(k)}$ and $\mathbf{h}_{u}^{(i)}$, while eliminating some inconsistent and noisy information.

\subsubsection{Optimization and final representation} 
In our collaborative training objective, the goal is to reconstruct the node-to-node interactions in ${\bf A}$ by using $y_{vu}$ and the node-to-attribute-category interactions in ${\bf X}^*$ by using $y_{v\delta_j}$.  The corresponding objective function is defined as follows.
\begin{equation}
\begin{aligned}
\mathcal{L}=&-\sum\nolimits_{v\in\mathcal{V}}\sum\nolimits_{z\in\mathcal{N}_v}\log\frac{\exp(\mathbf{y}_{vz})}{\sum_{z'\in\mathcal{V}\cup \mathcal{U}}\exp(\mathbf{y}_{vz'})} \\
&-\sum\nolimits_{\delta_j\in\mathcal{U}}\sum\nolimits_{u\in\mathcal{N}_{\delta_j}}\log\frac{\exp(\mathbf{y}_{{\delta_j}u})}{\sum_{u'\in\mathcal{V}\cup \mathcal{U}}\exp(\mathbf{y}_{{\delta_j}u^{'}})}.
\end{aligned}
\label{eq_lossconn}
\end{equation}
Eq.~\eqref{eq_lossconn} is usually intractable in practice, because the sum operation of the denominator is computationally prohibitive. Therefore, we employ the negative sampling~\cite{tang2015line,hamilton2017inductive} strategy to accelerate the optimization. 

After we have trained the model CONN by using the collaborative objective in Eq.~\eqref{eq_lossconn}, we need to define the final embedding representation of nodes to perform the downstream tasks. 
For link prediction task, we directly predict the probability of linking based on $\mathbf{y}_{vu}$ for a node pair $(v,u)$. For node classification, we adopt the output from the last layer $\mathbf{H}^{(K)}$ as node embedding, in which off-the-shelf classification algorithms could directly use it to classify. 

\subsection{Comparison with Prior Work}
To the best of our knowledge, few efforts have been devoted to leverage node attributes to redefine the core components of embedding methods for graphs. We roughly divide them into two categories and analyze the difference below.

\noindent\textbf{Random-walk based approach}. Random-walk based methods focus on conducting truncated random walks to generate node sequences, and then applying Skip-gram algorithm to learn node representations based on the sequences. This approach is initially not applicable for node attributes. ANRL~\cite{zhang2018anrl} addresses this issue by modifying the loss function of Skip-gram to depend on node attributes. 
Featwalk~\cite{huang2019large} suggests to conduct an attribute-aware random walks to inject node attributes in the random walk generation process. Compared with these methods, our model belongs to graph neural network approach that takes advantage of graph convolutional networks to explicitly model local graph structure.

\noindent\textbf{Graph neural network based approach}. This line of methods focuses on exploiting graph convolutional networks~\cite{kipf2016semi} to model the local subgraph of an anchor node for node representation. It is natural to cope with attributed graphs by using node attributes as initial node features in the first layer. However, such approach is rather limited in exploiting node attributes as they are excluded from the two crucial components of GCN, i.e., neighborhood message propagation and the training objective. CAN~\cite{meng2019co} attempts to alleviate this problem by jointly reconstructing node attributes and network structure under variational autoencoder. In contrast, our model focuses on exploiting node attributes to impact the two building blocks jointly. 

{Besides, we also want to remark that the utilization of node attributes in our work is different from skip connection trick~\cite{li2020autograph}. Skip connection targets to skip some higher-order neighbors in deeper layers by looking back initial features, but our focus is to redefine the neighborhood set of nodes in different orders by regarding attribute categories as "neighbor". Namely, our model introduces additional node-to-attribute-category relations in the message aggregation process of each layer. In a summary, skip connection is complementary with our model and can be added to our GNN backbone to avoid over-smoothing.} 

\noindent\textbf{Heterogeneous network embedding.} Thanks to our proposal in constructing the augmented network between nodes and attribute categories in Section~\ref{augmented_network}, we can also regard the resultant embedding task as a special heterogeneous network embedding (HNE) problem~\cite{yang2020heterogeneous,zhang2019heterogeneous,wang2019heterogeneous,jin2021heterogeneous}, where we have two object types (node and attribute category) and two relation types (node-to-node and node-to-attribute-category). Under this principle, state-of-the-art heterogeneous models might be applied. However, we found that such approach may make the learning task unnecessarily complex, since HNE will over-emphasize the heterogeneity of attributed networks. This may be suboptimal because attributed networks are not truly "heterogeneous" graphs. Moreover, it's nontrivial to train the model well, because no features are available for the second object--attribute-category. Also, it's hard to control the importance of two relations or node types in a mixed manner, such as mixed random-walk~\cite{zhang2019heterogeneous}. In contrast, by regarding the augmented network as homogeneous graph in our setting, the problem itself is substantially simplified and arbitrary standard GNNs architectures can be used in the plug-and-play fashion, equipping our proposal with broader applicability and practicability. 

To summarize, we propose a new alternative approach for GNNs based attributed network embedding by regarding attribute category as additional nodes and converting attributed network into an augmented graph. The augmented graph offers flexible information diffusion between nodes and attribute categories without information loss. To effectively learn node representations from the new graph, we develop two tailored designs: collaborative aggregation and cross-correlation mechanism, where the former helps to explicitly control the information propagation between nodes and node attributes in convolution, while the latter improves model's reconstruction capacity using multi-granularity features. Although our proposal seems general and is applicable for both homogeneous GNNs and heterogeneous GNNs, we empirically found that it's nontrivial to learn node representations based on heterogeneous GNNs (See discussion in~\ref{heterogenous_varaint}). Therefore, we focus on the simple case and leave attributed network embedding based on heterogeneous GNNs as the future work.     

\section{Experiments}\label{sec:experiment}
We analyze the effectiveness of CONN on multiple real-world datasets with various scales and types. Specifically, our evaluation centers around three questions.
\begin{itemize}
    \item \textbf{Q1:} Compared with the state-of-the-art embedding methods, can CONN achieve better performance in terms of node classification and link prediction tasks?
    
    \item \textbf{Q2:} There are three crucial components, i.e., graph mixing convolutional layer, graph correlation layer, and collaborative optimization, in CONN, how much does each component contribute? 
    
    \item \textbf{Q3:} What are the impacts of hyperparameters: the trade-off parameter $\alpha$ and embedding dimension $d$, on CONN?

\end{itemize}

\begin{table}[t]
\centering
  \caption{Statistics of the datasets.}
  \begin{tabular}{c| c |c |c | c}
   \toprule
     &$|\mathcal{V}|$ &\# Edge &$|\mathcal{U}|$ &$\# Label$\\
     \hline
    Pubmed &$19$,$717$ &$44$,$338$ &$500$ & $3$\\
    ACM &$48$,$579$  &$119$,$974$ &$10$,$000$ &$9$\\
    BlogCatalog &$5$,$196$  &$171$,$743$ &$8$,$189$ &$6$\\
    ogbn-arxiv &169,343  &1,166,243 &128 &40\\
    Reddit &$232$,$965$   &$11$,$606$,$919$ & $602$  & $41$  \\
    ogbl-collab &$235,868$   &$1,285,465$ & $128$  & $-$  \\
 \bottomrule
\end{tabular}
\vspace{-8pt}
\label{table2}
\end{table}

\subsection{Datasets}
We conduct experiments on six publicly available attributed networks of various scales and types. Their statistical information is summarized in Table~\ref{table2}. 

\textbf{Pumbed}~\cite{sen2008collective}. It is the biggest benchmark citation network used in~\cite{kipf2016semi}. Nodes correspond to documents and edges correspond to citations. Each node has a bag-of-words feature vector according to the paper abstract. Labels are defined as the academic topics.  

\begin{table*}[htbp]
\centering
  \caption{Node classification performance. }
  \begin{normalsize}
\setlength{\tabcolsep}{3.5pt}
  {
    \begin{tabular}{l cc cc cc cc cc}
    \toprule
     \multirow{2}{*}{Method} &\multicolumn{2}{c}{Pubmed} &\multicolumn{2}{c}{ACM} &\multicolumn{2}{c}{BlogCatalog} &\multicolumn{2}{c}{Reddit} &\multicolumn{2}{c}{ogbn-arxiv}\\
    \cmidrule(r){2-3} \cmidrule(r){4-5} \cmidrule(r){6-7} \cmidrule(r){8-9} \cmidrule(r){10-11}
     &F1-micro & F1-macro &F1-micro & F1-macro
    &F1-micro & F1-macro &F1-micro & F1-macro &F1-micro & F1-macro\\
    
     \cmidrule(r){2-3} \cmidrule(r){4-5} \cmidrule(r){6-7} \cmidrule(r){8-9}\cmidrule(r){10-11} 
    GAE &$0.825 $ &$0.819 $ &$0.710 $ &$0.611 $ &$0.636$ &$0.632 $ & $0.624 $ &$0.467 $ &$0.614$ &$0.401$\\
    ARGE &$0.837$ &$0.833$ &$0.736$ &$0.674$ &$0.606$ &$0.601$ &$0.645$ &$0.601$ &$0.633$ &$0.423$ \\
    DSGC &$0.844$ &$0.841$ &$0.759$ &$0.705$ &$0.644$ &$0.639$ &$0.693$ &$0.611$ &$0.643$ &$0.426$ \\
    DGI &$0.857$ &$0.856 $ &$0.768$ &$0.709 $ &$0.753 $ &$0.750 $ &$0.605 $&$0.418 $ &$0.646$ &$0.415$\\
    GIC &$0.859$ &$0.857$ &$0.754$ &$0.698$ &$0.773$ &$0.780$ &$0.622$&$0.433$ &$0.628$ &$0.427$\\
    GCA &$\bf{0.864}$ &$\bf{0.860}$ &$0.757$ &$0.702$ &$0.815$ &$0.836$ &$0.752$ &$0.639$ &$0.655$ &$0.431$ \\
    FeatWalk &$0.843 $ &$0.843 $ &$0.760 $ &$0.703 $ &$0.935 $ &$0.934 $ &$0.663 $ &$0.503 $ &$0.637$ &$0.424$ \\
    CAN &$0.841$ &$0.835 $ &$0.721 $ &$0.657 $ &$0.652 $&$0.648 $ &$0.820 $ &$0.726 $ &$0.650$ &$0.428$\\
    \midrule
    CONN &$\bf{0.866}$ &$\bf{0.862 }$ &$\bf{0.775 }$ &$\bf{0.723 }$ &$\bf{0.945 }$ &$\bf{0.944 }$ &$\bf{0.913}$ &$\bf{0.879 }$ &$\bf{0.674}$ &$\bf 0.456$\\
  \bottomrule
\end{tabular}}
\end{normalsize}
\label{table_classify}
\end{table*}
\textbf{ACM}~\cite{tang2008arnetminer}. It is a large-scale citation network consisting of 48,579 papers published in ACM. Words in the paper abstracts are adopted as node attributes based on the bag-of-words model. Citation links are treated as edges. Each paper is published under a specific area which serves as label for classification.

\textbf{BlogCatalog}~\cite{huang2017label}. It is a social network collected from a blog community. Nodes are web users and edges indicate the user interactions. Node attributes denote the keywords of their blogs.
Each user could register his/her blogs into six different predefined classes, which are considered as class labels for node classification.


\textbf{Reddit}~\cite{hamilton2017inductive}. It is another social network dataset collected from the online discussion forum-Reddit. Nodes correspond to the Reddit posts and edges represent the co-comment relationships. 
The posts are preprocessed into 602-dimensional feature vectors via Glove CommonCrawl world embedding~\cite{pennington2014glove}. Hence, node attributes refer to the 602 latent dimensions.   
We use the communities or `subreddit' that the post belongs to a target label. 

{\textbf{ogbl-collab}~\cite{wang2020microsoft}. } It is a challenging author collaboration network from KDD Cup 2021. Each node is an author and edges indicate the collaboration between authors. All nodes come with 128-dimensional features, obtained by averaging the word embeddings of papers that are published by the authors. It is widely used to conduct link prediction task. 

{\textbf{ogbn-arxiv}~\cite{wang2020microsoft}.} It is a large-scale paper citation network of arXiv papers from KDD Cup 2021. Each node is an arXiv paper and the edge indicates that one paper cites another one. Each paper is represented by a 128-dimensional feature vector obtained by averaging the embeddings of words in its title and abstract. The target is to predict 40 subject areas.

\subsection{Baseline Methods}
To validate the effectiveness of CONN, we include four categories of unsupervised baselines as follows. First, to study why we need tailored framework to incorporate node attributes into GCN-architectures, we compare with vanilla GCN methods GAE~\cite{kipf2016variational}. Second, to investigate how effective is CONN compared with other tailored solutions, we include two recent works, i.e., CAN~\cite{meng2019co} and FeatWalk~\cite{huang2019large}. Third, to have a comprehensive evaluation with state-of-the-art unsupervised models, we include three popular self-supervised learning based GNN methods, DGI~\cite{velickovic2019deep}, GIC~\cite{mavromatis2021graph}, and GCA~\cite{zhu2021graph}. Note that, other non-GCN based embedding methods are not included, i.e., DeepWalk~\cite{perozzi2014deepwalk}, LINE~\cite{tang2015line} and ANRL~\cite{zhang2018anrl}, since they are outperformed by CAN and FeatWalk in their experiments~\cite{meng2019co,huang2019large}.
Besides, other classical GNNs architectures, i.e., GAT~\cite{velivckovic2017graph}, SGC~\cite{wu2019simplifying}, and APPNP~\cite{klicpera2019predict}, are not included for comparison, since they are initially dedicated for supervised learning while we focus on unsupervised representation learning. 
\begin{itemize}
\item\textbf{GAE}~\cite{kipf2016variational}. It learns node embeddings by reconstructing the network structure under the autoencoder approach. Specifically, it employs graph convolutional network to encode a subgraph into latent space. 

\item {\textbf{ARGE}~\cite{pan2018adversarially}.It is an adversarially regularized GAE. We do not consider the variational version since the two variants perform quick similar in most cases.}

\item \textbf{DGI}~\cite{velickovic2019deep}. It learns node embeddings by maximizing the mutual information between the local patch representation and the global graph representation. 

\item \textbf{GIC}~\cite{mavromatis2021graph}. It updates DGI by leveraging cluster-level node representation for unsupervised representation learning. 

\item\textbf{GCA}~\cite{zhu2021graph}. It learns node representations by minimizing the contrastive loss between the original graph and its augmented forms. 

\item \textbf{CAN}~\cite{meng2019co}. It learns node embeddings by reconstructing both the network structure and attribute matrix under the variational autoencoder framework. 

\item \textbf{FeatWalk}~\cite{huang2019large}. It advances vanilla random-walk based methods via introducing an attribute-enhanced random walk strategy, which helps to generate diversified random walks for representation learning.  

\item {\textbf{DSGC}~\cite{li2021dimensionwise}. It defines a $k$-NN graph based on node features and then uses it as an attribute-aware graph filter for network embedding}.
\end{itemize}

Besides, we also introduce three variants to validate the effectiveness of core components in CONN.

\begin{itemize}
\item\textbf{CONN-gcn}. It replaces the graph mixing convolutional layer with vanilla graph convolutional layer, to verify the effectiveness of modeling mixed neighbors, i.e., nodes and attributes. 

\item\textbf{CONN-inner}. It excludes the graph correlation layer and utilizes the inner-product to estimate the similarity between two nodes based on their last layer representations similar to~\cite{hamilton2017inductive,kipf2016variational}. We use it to verify the usefulness of the proposed correlation layer.  

\item \textbf{CONN-ncoll}. It only considers the node-to-node interactions in the objective function. This variant is used to certify the contribution of jointly optimizing the node-to-node and node-to-attribute interactions.   
\end{itemize}

\subsection{Experimental Settings}
We follow the common protocol~\cite{kipf2016variational, meng2019co} to evaluate the performance of CONN. The effectiveness of the learned latent representations is evaluated over two popular downstream tasks, i.e., link prediction and node classification. For link prediction task, we randomly split 85\%, 10\% and 5\% edges in the network to form training, testing and validation sets similar to~\cite{kipf2016variational}. The link prediction task aims to estimate whether a missing edge in the network should be connected or not, based on its embedding representations. 
The performance is measured by two standard metrics, i.e., area under the ROC curve (AUC) and average precision (AP) scores. 

For node classification task, it targets to classify a new instance into one or multiple categories, based on the obtained node representations and the trained classifier. Specifically, we apply 5-fold cross-validation on all datasets to construct the training and test sets. To perform classification, we build an SVM classifier based on scikit-learn package and train the classifier based on the nodes in the training group and corresponding labels. Then we apply the learned classifier to predict the labels of instances in the test groups. The averaged results of five-fold is reported. 

We use the official released codes of baselines for experiments and use the validation set to tune their parameters. For our method, we train CONN for 100 epochs using Adam optimizer with learning rate 0.01 and early stopping with a patience of 20 epochs. If it is not specified, $d$ is set as 128, {$K=2$}, and $\alpha$ equals to 0.2 and 0.8 for node classification and link prediction tasks, respectively. For continuous datasets (Reddit, ogbn-arxiv, and ogbl-collab), we use top-$50$ values in each row of $\mathbf{X}$ to construct the bipartite graph by default. The source code of CONN is available at~\href{https://github.com/Qiaoyut/CONN}{https://github.com/Qiaoyut/CONN}.

\begin{table*}[htbp]
\centering
\caption{Link prediction results.}
 \begin{normalsize}
\setlength{\tabcolsep}{5pt}
{
\begin{tabular}{l cc cc cc cc cc cc cc}
\toprule
  &\multicolumn{2}{c}{Pubmed}
 &\multicolumn{2}{c}{ACM}
 &\multicolumn{2}{c}{BlogCatalog}
 &\multicolumn{2}{c}{Reddit}
  &\multicolumn{2}{c}{ogbl-collab}\\
\cmidrule(r){2-3} \cmidrule(r){4-5} \cmidrule(r){6-7} \cmidrule(r){8-9} \cmidrule(r){10-11} 
 &AUC &AP &AUC &AP
&AUC &AP &AUC &AP &AUC &AP\\

\cmidrule(r){1-3} \cmidrule(r){4-5} \cmidrule(r){6-7} \cmidrule(r){8-9}  \cmidrule(r){10-11} 
GAE &$0.920 $ &$0.911 $ &$0.957 $ &$0.956 $ &$0.824 $&$0.822 $&$0.578 $ &$0.565 $ &$0.821$ &$0.737$\\
ARGE &$0.968$ &$0.971$ &$0.964$ &$0.969$ &$0.755$ &$0.723$ &$0.603$ &$0.597$ &$0.847$ &$0.814$ \\
DSGC &$0.964$ &$0.970$ &$0.961$ &$0.965$ &$0.788$ &$0.774$ &$0.593$ &$0.586$ &$0.842$ &$0.808$ \\
DGI &$0.942 $ &$0.927 $ &$0.582$ &$0.644 $ &$0.733 $&$0.737 $&$0.594 $ &$0.586 $ &$0.818$ &$0.728$\\
GIC &$0.937$ &$0.935$ &$0.674$ &$0.775$ &$0.748$ &$0.745$ &$0.693$ &$0.677$ & $0.892$ &$0.804$\\
GCA &$0.955$ &$0.956$ &$0.756$ &$0.820$ &$0.826$ &$0.808$ &$0.717$ &$0.741$ &$0.886$ &$0.833$\\
FeatWalk &$0.940 $&$0.941$&$0.963$&$0.963$&$0.617$&$0.617$&$0.852$ &$0.870 $ &$0.828$ &$0.797$\\
CAN &$0.980$ &$0.977 $ &$0.896 $ &$0.899 $ &$0.837 $ &$0.837 $&$0.909 $ &$0.903 $ &$0.901$ &$0.886$\\
CONN  &$\bf{0.994 }$ &$\bf{0.993 }$ &$\bf{0.986 }$ &$\bf{0.987 }$ &$\bf{0.918}$ &$\bf{0.906 }$ &$\bf{0.986}$ &$\bf{0.985}$ &$\bf 0.930$ &$\bf 0.912$\\
\bottomrule
\end{tabular}}
 \end{normalsize}
\label{table_lp}
\end{table*}

\begin{table*}[t]
\caption{Ablation study of CONN on node classification task.}
\label{table_ablation}
\begin{center}
\begin{small}
\setlength{\tabcolsep}{5pt}
\begin{tabular}{lccccc}
\toprule
 & &CONN-gcn  & CONN-inner &CONN-ncoll & CONN \\
\midrule
\multirow{5}{*}{\begin{minipage}{0.42in}F1-micro\end{minipage}}
&Pubmed &$0.795$ &$0.701$ &$0.851$ &$\bf{0.866}$\\
&ACM &$0.704$ &$0.643$ &$0.756$ &$\bf 0.775$\\
&BlogCatalog &$0.745$ &$0.794$ &$0.900$ &$\bf 0.945$\\
&Reddit &$0.891$ &$0.571$ &$0.891$ &$\bf 0.913$\\
&ogbn-arxiv &$0.614$ &$0.553$ &$0.635$ &$\bf 0.674$\\
\midrule
\multirow{5}{*}{\begin{minipage}{0.42in}F1-macro\end{minipage}}
&Pubmed &$0.795$ &$0.708$ &$0.846$ &$\bf 0.862$\\
&ACM &$0.633$ &$0.551$ &$0.694$ &$\bf 0.723$\\
&BlogCatalog &$0.740$ &$0.787$ &$0.898$ &$\bf 0.944$\\
&Reddit &$0.849$ &$0.464$ &$0.844$ &$\bf 0.879$\\
&ogbn-arxiv &$0.408$ &$0.388$ &$0.415$ &$\bf0.456$\\
\bottomrule
\end{tabular}
\end{small}
\end{center}
\vspace{-10pt}
\end{table*}

\subsection{Node Classification}
We start to evaluate the performance of CONN in node classification task (\textbf{Q1}). Table~\ref{table_classify} summarizes the results on five datasets in terms of micro-average and macro-average scores. 

From Table~\ref{table_classify}, we observe that CONN performs consistently better than other baselines across two evaluation metrics on datasets with categorical attributes (PubMed, ACM, and BlogCatalog) and continuous feature embeddings (Reddit  and ogbn-arxiv). It demonstrates the effectiveness of CONN. To be specific, compared with vanilla GCN variants, CONN improves 48.6\% and 9.8\% over GAE on BlogCatalog and ogbn-arxiv datasets in terms of micro-average score, respectively. {CONN improves 46.7\% and 4.8\% over ARGE on BlogCatalog and ogbn-arxiv datasets under the macro-average score, respectively.} This improvement validates the necessarity to design tailored  GCN architecture for attributed networks. CAN is also proposed to model on node attributes, but it loses to CONN significantly on five scenarios. The major difference between them is that CAN targets to jointly reconstruct node attributes and network under auto-encoder framework, while CONN aims to revise GCN architecture by explicitly leveraging node attributes to guide message propagation. This comparison certifies that a tailored GCN solution for attributed networks is more promising and effective. FeatWalk incorporates node attributes to random-walk based models, but it is outperformed by CONN in all cases. Taking Reddit dataset for example, CONN improves FeatWalk over 37.7\%. It is reasonable since our model can leverage structure information for node embedding while random-walk based methods fail to use that. {Although DSGC also constructs an attribute-aware graph for network embedding, it loses to CONN in almost all cases. This is because DSGC requires to construct a $k$-NN graph based on node features, which may delete a lot of important attribute information. }

Another promising observation is that CONN performs significantly better than state-of-the-art self-supervised competitors (DGI, GIC and GCA) in general. This results indicates the effectiveness of reconstructing original network structure for unsupervised representation learning. Besides, the performance gap between CONN and FeatWalk increases on continuous-value datasets (reddit and orgn-arxiv). This is mainly because FeatWalk reduces to random choice among all attribute categories as the sampling graph is fully-connected, which damages the quality of random walks.  



\begin{figure*}[htbp]
\centering
\subfigure[Trade-off parameter $\alpha$]{\!\!\!
    \begin{minipage}[t]{0.33\linewidth}
        \centering
        \includegraphics[width=5cm, height=3.3cm]{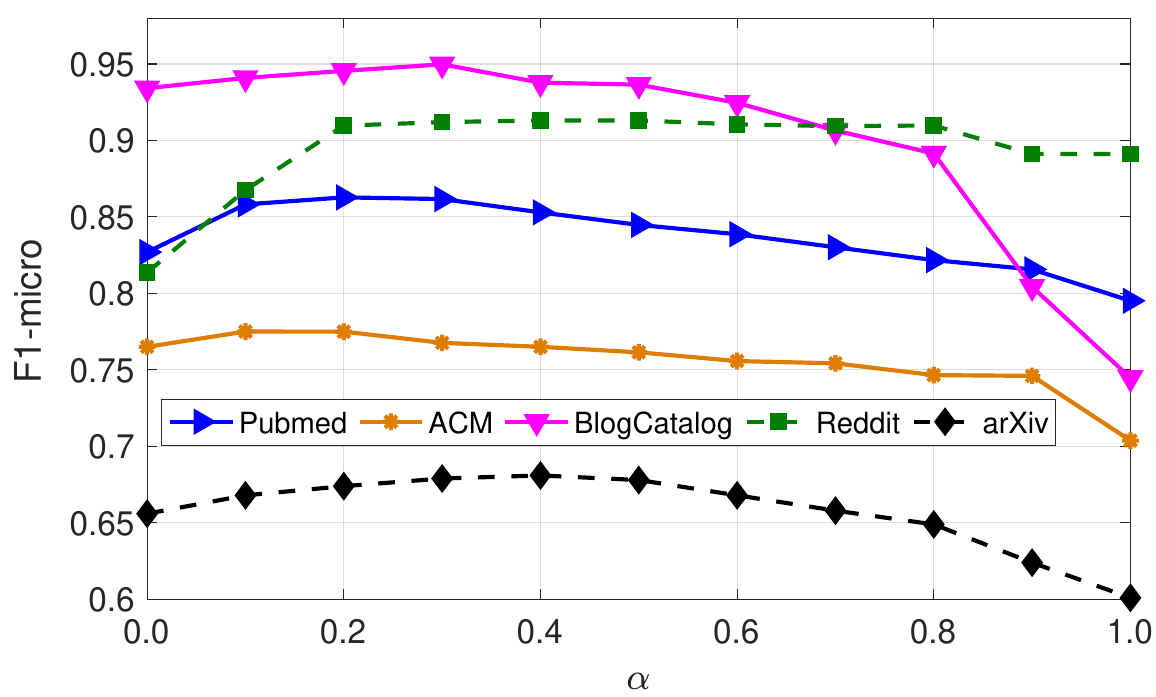}\\
        \vspace{0.02cm}
    \end{minipage}%
}\!\!\!\!\!\!
\subfigure[The number of layers $K$]{
    \begin{minipage}[t]{0.33\linewidth}
        \centering
        \includegraphics[width=5.5cm,height=3.3cm]{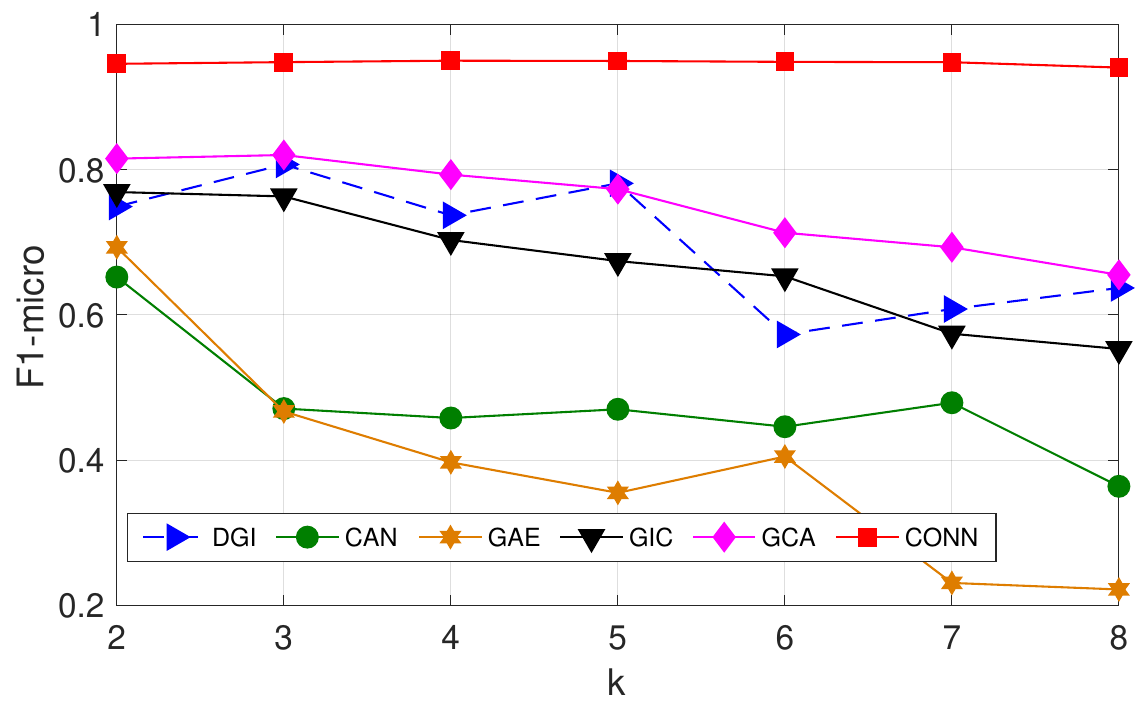}\\
        \vspace{0.02cm}
    \end{minipage}%
}\!\!\!\!\!\!
\subfigure[Embedding dimension $d$]{
    \begin{minipage}[t]{0.33\linewidth}
        \centering
        \includegraphics[width=6.4cm, height=3.4cm]{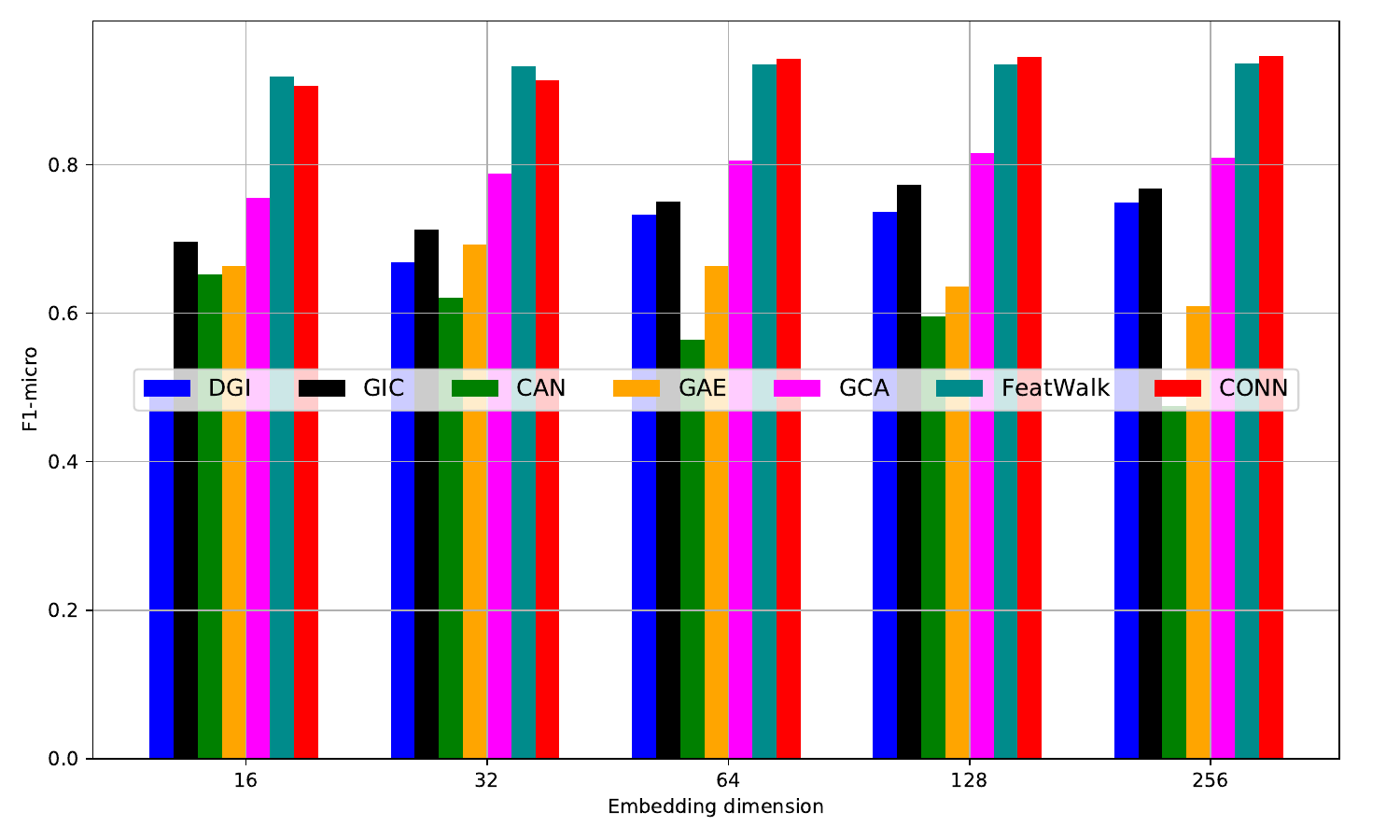}\\ 
        \vspace{0.02cm}
    \end{minipage}%
}
\centering
\caption{Hyper-parameter analysis of CONN.}
\label{figure_para}
\end{figure*}

\subsection{Link Prediction}
We now evaluate the performance of CONN in terms of link prediction task (\textbf{Q1}). Since DGI, GIC, GCA, and FeatWalk are not originally tested on this task, we delete the test edges from the adjacent matrix and use the resulting adjacent matrix to train them. Then, we estimate the similarity scores for test edges based on the inner product of corresponding node representations similar to~\cite{meng2019co}. Table~\ref{table_lp} reports the results on three median size (PubMed, ACM, and BlogCatalog) and two large-scale (Reddit and ogbl-collab) datasets in terms of AUC and AP scores.

From the table, we can see that CONN performs significantly better than other baselines. {Specifically, it improves 11.4\%, 21.5\%, 16.5\%, 25.2\%, 22.7\%, 11.1\%, 48.8\%, and 9.7\% over GAE, ARGE, DSGC, DGI, GIC, GCA, FeatWalk, and CAN on BlogCatalog in terms of AUC value, respectively.} CAN loses to FeatWalk on node classification task in most cases, but outperforms FeatWalk on predicting missing edges. It indicates the importance of capturing structure information for link prediction. Both CAN and CONN target to leverage node attributes for node embedding, but CONN outperforms CAN with a great margin in all cases. The main difference is that CAN focuses on using node attributes to enrich the objective function, with the hope to enhance GCN encoder optimization. In contrast, our model directly merges useful node attributes into the GCN building blocks, such that a more powerful GCN encoder could be explicitly achieved. Although DGI, GIC, and GCA are trained based on the advanced contrastive loss function, our model performs substantially better than them with a wide margin. These results indicate the insufficiency of existing GCN-based architectures in exploiting useful node attributes. Based on these observations, we believe that our proposed framework is more suitable for the link prediction task.

\subsection{Ablation Study}
\label{ablation_study}
We now investigate the second question (\textbf{Q2}), i.e., how much could CONN's three major components, i.e., graph mixing convolutional layer, graph correlation layer, and collaborative optimization contribute? Three variants, i.e., CONN-gcn, CONN-inner, and CONN-ncoll, that are introduced at the beginning, are used for this ablation study. Table~\ref{table_ablation} records the node classification performance on five datasets in terms of micro-and macro-average scores.

Based on Table~\ref{table_ablation}, we have three major observations. First, without node attributes to guide message propagation in the graph convolution process, the performance of CONN-gcn decreases. It validates the effectiveness of explicitly incorporating node attributes into GCN architecture. Second, without the graph correlation layer, CONN-inner loses to CONN with great margin. For instance, CONN achieves 23.5\% improvements than CONN-inner on Pubmed in terms of micro-average. The major difference between CONN-inner and CONN is that the former one adopts simple inner product to estimate edge similarity, while CONN devises deep correlation layer, which is capable of capturing the complex correlations between two nodes. This comparison verifies the effectiveness of the proposed graph correlation layer. Third, CONN performs slightly better than CONN-ncoll across five datasets. It indicates that jointly optimizing node-to-node and node-to-attribute interactions is beneficial. Given that CONN outperforms three variants, it verifies that we propose a principled framework to reinforce the reciprocal effects among the three components. 

\subsection{Parameter Sensitivity Analysis}
We now study the impact of parameters $\alpha$ and embedding dimension $d$ over CONN on BlogCatalog (\textbf{Q3}). $\alpha$ controls the importance of node-to-node interaction and node-to-attribute interaction for message passing. We plot the performance of CONN when $\alpha$ varies from $0$ to $1$ with step size 0.1 on Figure~\ref{figure_para}-a. From the results, we observe that the performance of CONN increases as $\alpha$ increases from 0.0 to 0.2 on five datasets. CONN obtains the best results when $\alpha=0.2$ in general. Notice that, when $\alpha=0$, CONN only utilizes the node attribute interactions for graph convolution, and the network structure is excluded. When $\alpha=1$, node-to-attribute interactions are not considered. CONN reduces to the vanilla GCN, except that both node attribute and node interactions are jointly optimized.

$K$ is the number of layers for GCN backbones. 
Large $K$ means that high-order neighbors are included. Since FeatWalk has no such parameter, we omit it for comparison. We vary $K$ from 2 to 8, and the performance on BlogCatalog is shown in Figure~\ref{figure_para}-b. From the results, we observe that our model CONN consistently outperforms all baselines on the different number of layers. Another interesting observation is that when $K$ varies from 2 to 8, the performance of CAN, GAE, GIC, GCA, and DGI decreases while CONN performs stable. It validates the superiority of our model in capturing high-order dependencies. {It is worth noting that although CONN achieves relatively stable results than standard GNN models in Figure 2 (b) when K is small, e.g., 8, we do observe an obvious performance drop when K is larger, i.e., 15. Therefore, by modeling attribute-aware relationships, our method can alleviate the over-smoothing problem when the GNN model is relatively shallow to some extent. Still, tailored efforts are required to thoroughly tackle the over-smoothing issues as done in~\cite{chen2020simple}.   }

In CONN, the dimension $d$ presents the output latent space for downstream applications, i.e., node classification and link prediction. We search it from $\{16, 32, 64, 128, 256\}$, and the performance of all methods are depicted in Figure~\ref{figure_para}-c. We observe that different methods have different optimal $d$, and our model CONN could achieve satisfactory performance when $d=128$. Similar observations are made on other datasets. To make it fair, we tune the best embedding dimension for all methods and report their best results.

\begin{figure}[htbp]
\centering
\subfigure[ogbl-collab]{\!\!\!
    \begin{minipage}[t]{0.5\linewidth}
        \centering
        \includegraphics[width=4.5cm, height=3.5cm]{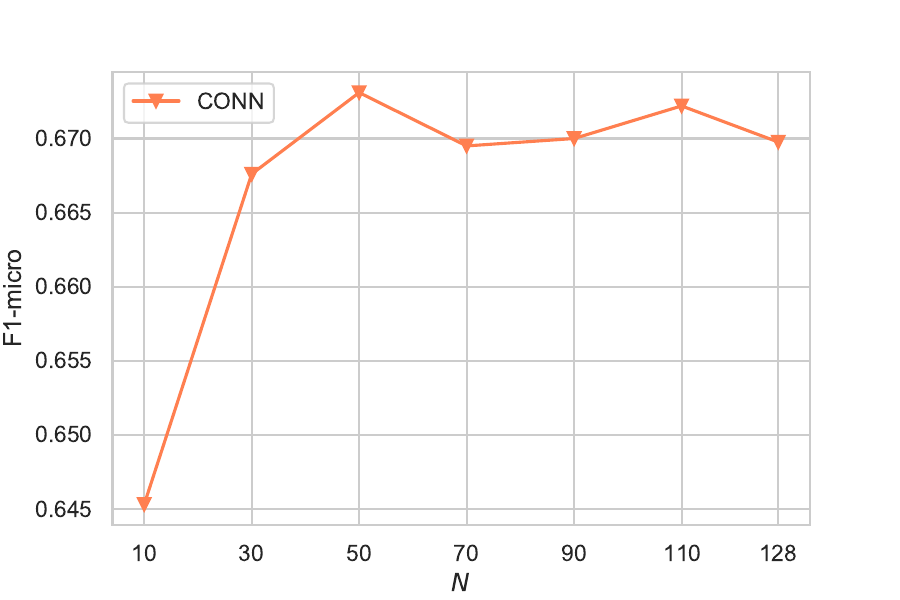}\\
        \vspace{0.02cm}
    \end{minipage}%
}\!\!\!\!\!\!
\subfigure[Reddit]{
    \begin{minipage}[t]{0.5\linewidth}
        \centering
        \includegraphics[width=4.5cm,height=3.5cm]{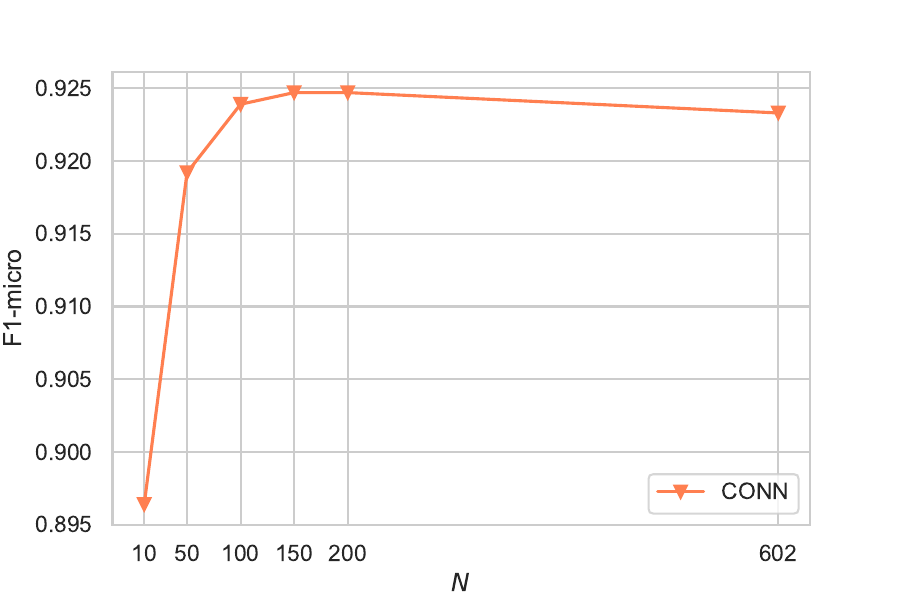}\\
        \vspace{0.02cm}
    \end{minipage}%
}
\centering
\caption{CONN performance \textit{w.r.t.} different top-$N$ values.}
\label{figure_topn}
\end{figure}

\subsection{Further Analysis}
\label{sec:further}
With performance comparison with SOTA methods completed, we delve into our proposal for even more insights.

\subsubsection{\textbf{Constructing a sparse bipartite graph from continuous features is a good trade-off.}} The first question we would like to answer is: what is the impact of top-$N$ values for graphs with continuous features? Figure~\ref{figure_topn} shows the results of our model \textit{w.r.t.} different $N$ values on two graphs (Reddit and ogbl-collab) with embedding vectors (e.g., continuous values) as node attributes.  
From the figures, we can see that the performance of CONN increases with the increasing of $N$ values, and it achieves good results when $N$ equals the dimension of node attributes, i.e., 128 and 602 for collab and reddit, respectively. 

This observation sheds light on the following insights. (i) Top-$N$ strategy is a good way to handle continuous feature vectors in our model, since $N=50$ can already achieve satisfactory results in both cases. (ii) Furthermore, it indicates that our model can be used in high-dimensional data, since we can first adopt classical dimensional reduction techniques to reduce the dimension. 

To check the scalability of our model on high-dimensional data, we first adopt Deepwalk to obtain 128-d embedding vectors for nodes in ACM and BlogCataglog, and then use the resultant node features as input to train our model. We observe comparable results in two downstream tasks. Specifically, the AUC results for ACM and BlogCatalog on link prediction are 0.987 and 0.916; while the Micro-average results on node classification are 0.769 and 0.941.   

\subsubsection{\textbf{Modeling node attributes as a bipartite graph enhances the robustness of GNNs for representation learning.} } 
\begin{figure}[htbp]
\centering
\subfigure[BlogCatalog]{\!\!\!
    \begin{minipage}[t]{0.5\linewidth}
        \centering
        \includegraphics[width=4.5cm, height=3.3cm]{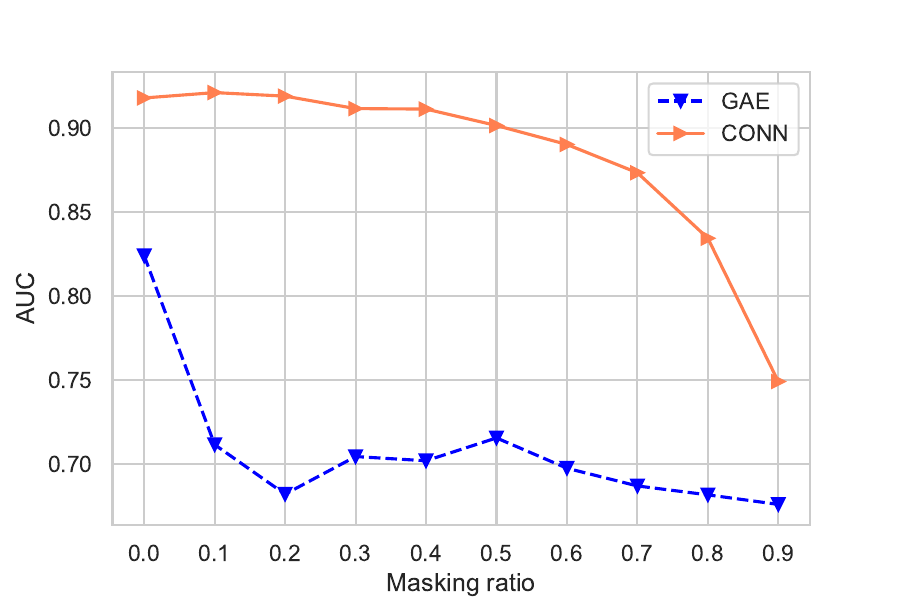}\\
    \end{minipage}%
}\!\!\!\!\!\!
\subfigure[Reddit]{
    \begin{minipage}[t]{0.5\linewidth}
        \centering
        \includegraphics[width=4.5cm,height=3.3cm]{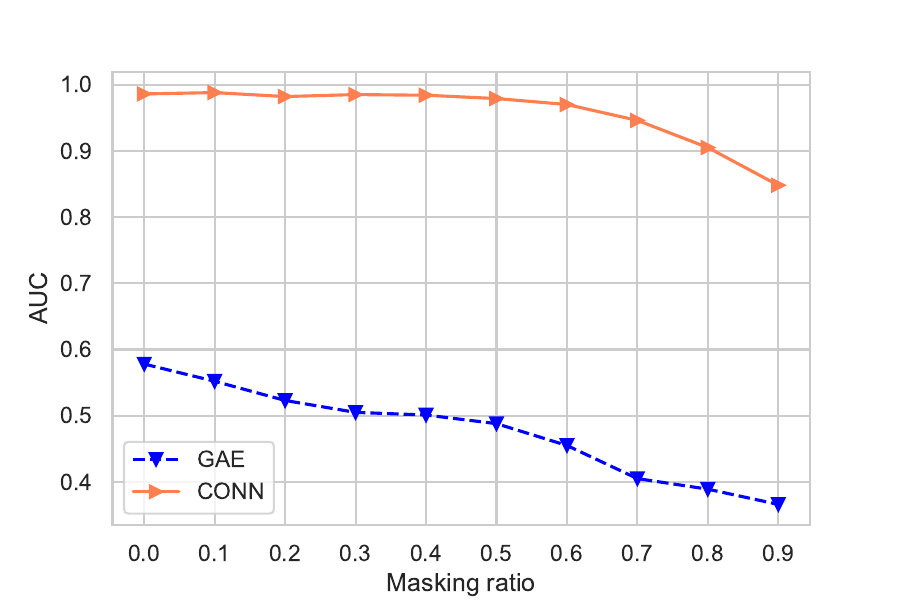}\\
    \end{minipage}%
}
\centering
\caption{Robustness analysis of CONN vs. GAE with different edge perturbation ratios.}
\label{figure_sens}
\end{figure}

Next, we explore whether modeling node attributes as an augmented graph can improve model's robustness towards noises, i.e., missing links.
We analyze the robustness of our model and the vanilla GNNs competitor (GAE) towards edge perturbation, i.e., randomly masking some edges with ratios, ranging from 0.0 to 0.9 with step size 0.1. We show the results in terms of link prediction in Figure~\ref{figure_sens}. Similar sensitive tendencies are observed in other datasets. 

The results in two figures show that our model performs relatively stable when the masking ratio is less than 50\% across two datasets, while the performance of GAE drops in general when the ratio increases. We believe this stable merit is attributed to the proposed augmented bipartite graph because it can replenish missing connections between nodes via using node categories as intermedia. 

\subsubsection{\textbf{Jointly reconstructing node attributes and node interactions does not impact the convergence speed.}} 
\begin{figure}[htbp]
\centering
\subfigure[BlogCatalog]{\!\!\!
    \begin{minipage}[t]{0.5\linewidth}
        \centering
        \includegraphics[width=4.5cm, height=3.5cm]{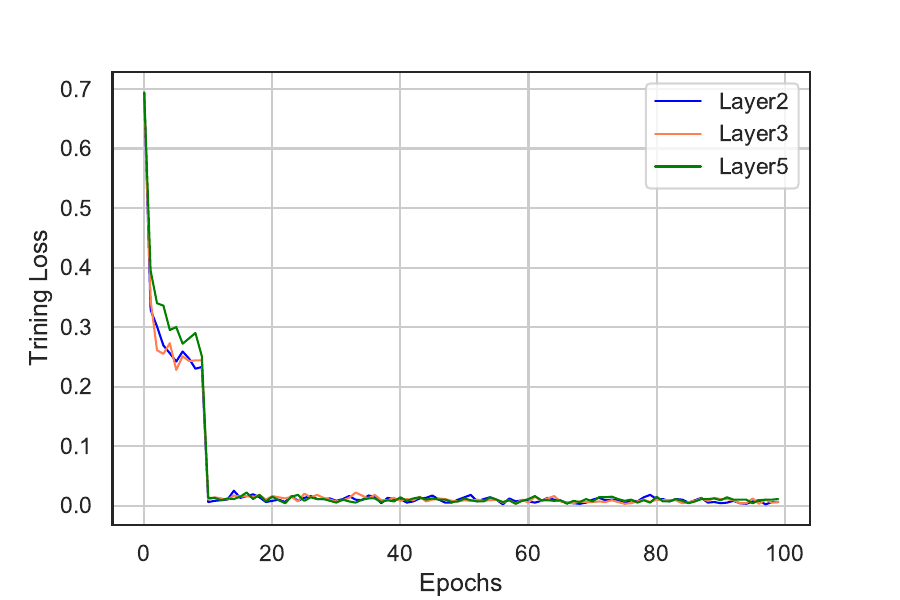}\\
    \end{minipage}%
}\!\!\!\!\!\!
\subfigure[Reddit]{
    \begin{minipage}[t]{0.5\linewidth}
        \centering
        \includegraphics[width=4.5cm,height=3.5cm]{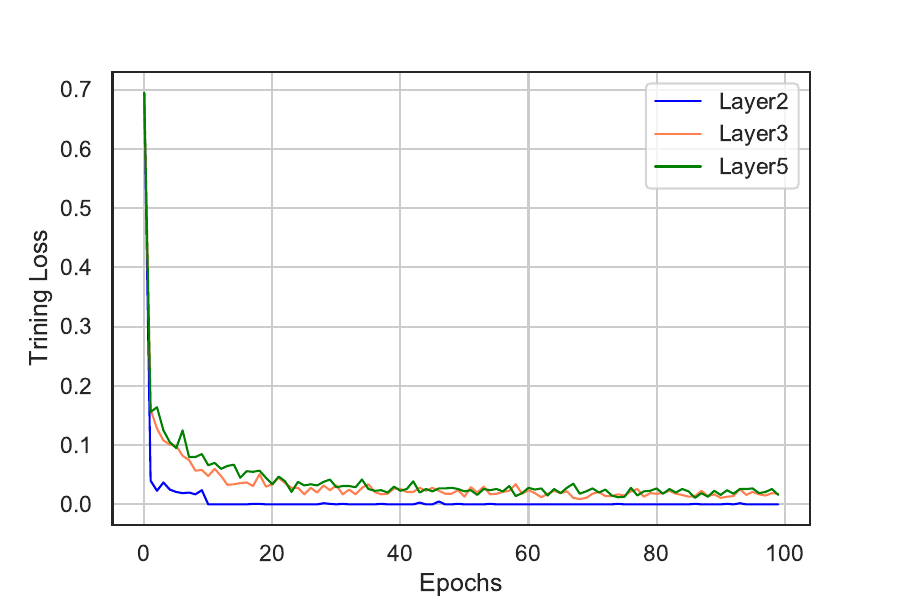}\\
    \end{minipage}%
}
\centering
\caption{Empirical training curves of CONN on two datasets with different GNN layers.}
\label{figure_loss}
\end{figure}

Moreover, we would like to examine the empirical convergence of our model. We show the training curves of CONN under different GNNs layers over two representative datasets (BlogCatalog and Reddit for categorical and continuous attributes, respectively) in Figure~\ref{figure_loss}. We can observe that the training loss decreases very fast in the first 10 epochs, and our model tends to converge within 50 epochs in general.


\begin{table}[t]
\caption{Performance of CONN vs. heterogeneous GNNs methods. "NC" stands for node classification results in terms of F1-micro metric; "LP" stands for link prediction results in terms of AUC.}
\label{table_hne}
\begin{center}
\begin{small}
\setlength{\tabcolsep}{5pt}
\begin{tabular}{lcccc}
\toprule
 & &HetGNN  &MAGNN & CONN \\
\midrule
\multirow{5}{*}{\begin{minipage}{0.42in}NC\end{minipage}}
&Pubmed &$0.836$ &$0.839$ &$\bf{0.866}$\\
&ACM &$0.724$ &$0.726$ &$\bf 0.775$\\
&BlogCatalog &$0.785$ &$0.816$ &$\bf 0.945$\\
&Reddit &$0.646$ &$0.659$ &$\bf 0.913$\\
&ogbn-arxiv &$0.625$ &$0.637$ &$\bf 0.674$\\
\midrule
\multirow{5}{*}{\begin{minipage}{0.42in}LP\end{minipage}}
&Pubmed &$0.936$ &$0.942$ &$\bf 0.994$\\
&ACM &$0.869$ &$0.877$ &$\bf 0.986$\\
&BlogCatalog &$0.808$ &$0.812$ &$\bf 0.918$\\
&Reddit &$0.646$ &$0.681$ &$\bf 0.986$\\
&ogbn-collab &$0.834$ &$0.856$ &$\bf0.930$\\
\bottomrule
\end{tabular}
\end{small}
\end{center}
\vspace{-10pt}
\end{table}

\subsubsection{\textbf{Regarding the augmented network as heterogeneous graph is suboptimal.}} {Finally, we investigate the generalization of our proposal in terms of heterogeneous GNNs. In particular, we want to explore whether the well-established heterogeneous GNNs efforts can be directly applied for attributed network embedding by using our proposed augmented network. To this end, we select two popular unsupervised heterogeneous GNN embedding methods as backbone: HetGNN~\cite{zhang2019heterogeneous} and MAGNN~\cite{fu2020magnn}. For a fair comparison, we initialize learnable embeddings for nodes and attribute categories similar to our model. Table~\ref{table_hne} reports the results on both node classification and link prediction tasks. }

{We observe a clear performance gap between our CONN and two heterogeneous GNNs methods (HetGNN and MAGNN) on all evaluation scenarios. Jointly considering the results in Table~\ref{table_classify} and~\ref{table_lp}, the performance of two heterogeneous variants ranks in the middle against all baselines. The possible explanation is that regarding augmented networks as a heterogeneous graph may over-emphasize the heterogeneity between node interactions and node attributes. It is hard to control the importance between them via either random walk~\cite{zhang2019heterogeneous} or meta-path~\cite{fu2020magnn}. These comparisons demonstrate our motivation to treat the augmented network as a homogeneous graph. We believe nontrivial efforts are needed to unleash the power of heterogeneous GNNs on attributed network embedding. }
\label{heterogenous_varaint} 

\section{Conclusion}\label{sec:experiment}
In this paper, we study the problem of unsupervised node representation learning on attributed graphs under graph neural networks (GNNs). Existing GNNs efforts mainly focus on exploiting topological structures, while only using node attributes as initial node presentations in the first layer. Here, we argue that such GNN architecture is suboptimal to modeling real-world attributed graphs, since node attributes are totally excluded from the key factors of GNNs, i.e., the message aggregation mechanism and the training objectives. To tackle this problem, we propose a novel collaborative graph neural network termed CONN. It allows node attributes to determine the message-passing process for neighborhood aggregation and enables the node-to-node and node-to-attribute-category interactions to be jointly recovered. Empirical results on node classification and link prediction tasks over social and citation graphs demonstrate the superiority of CONN against state-of-the-art embedding methods. Our future work is to explore its applicability for dynamic graphs and further improve the robustness of our model by integrating adversarial training. Moreover, we are interested in exploring similar ideas in recommendation scenarios~\cite{zhadreamshard}, such as sequential recommendation~\cite{tan2021sparse,tan2021dynamic} and session recommendation~\cite{zhou2021temporal,zhou2023interest}.

\section{Acknowledgement}
We thank the anomalous reviewers for the feedback. The work is, in part, supported by NSF (IIS-2224843 and IIS-1849085).


%





\ifCLASSOPTIONcaptionsoff
  \newpage
\fi



\normalem
\bibliographystyle{IEEEtran}
\bibliography{reference}
\end{document}